\documentclass[letterpaper, 10 pt, conference]{ieeeconf}  

\IEEEoverridecommandlockouts                              

\usepackage{booktabs}
\usepackage{multirow}
\usepackage{graphicx}
\usepackage{comment}
\usepackage{amsfonts}
\usepackage{caption}
\usepackage{subcaption}
\usepackage{dsfont}
\usepackage{amsmath}
\usepackage{color}
\usepackage{colortbl}
\definecolor{color}{rgb}{1.0, 0.92, 0.92}

\makeatletter
\let\NAT@parse\undefined
\makeatother
\usepackage{hyperref}
\usepackage{cleveref}

\title{\LARGE \bf
I-FailSense: Towards General Robotic Failure Detection with Vision-Language Models
}

\author{Clémence Grislain*, Hamed Rahimi*, Olivier Sigaud, Mohamed Chetouani\\
ISIR,  Sorbonne Université, CNRS, Paris, France\\
  \small\texttt{\{grislain, rahimi, sigaud, chetouani\}@isir.upmc.fr}
}

\begin{document}

\maketitle
\begingroup
\renewcommand{\thefootnote}{$^{*}$}
\footnotetext[1]{\small Equal contribution}
\endgroup
\thispagestyle{empty}
\pagestyle{empty}

\begin{abstract}

Language-conditioned robotic manipulation in open-world settings requires not only accurate task execution but also the ability to detect failures for robust deployment in real-world environments. Although recent advances in vision-language models (VLMs) have significantly improved the spatial reasoning and task-planning capabilities of robots, they remain limited in their ability to recognize their own failures. In particular, a critical yet underexplored challenge lies in detecting semantic misalignment errors, where the robot executes a task that is semantically meaningful but inconsistent with the given instruction. To address this, we propose a method for building datasets targeting Semantic Misalignment Failures detection, from existing language-conditioned manipulation datasets. We also present \textit{I-FailSense}, an open-source VLM framework with grounded arbitration designed specifically for failure detection. Our approach relies on post-training a base VLM, followed by training lightweight classification heads, called \textit{FS blocks}, attached to different internal layers of the VLM and whose predictions are aggregated using an ensembling mechanism. Experiments show that I-FailSense outperforms state-of-the-art VLMs, both comparable in size and larger, in detecting semantic misalignment errors. Notably, despite being trained only on semantic misalignment detection, I-FailSense generalizes to broader robotic failure categories and effectively transfers to other simulation environments and real-world with zero-shot or minimal post-training. 
The datasets and models are publicly released on HuggingFace (\href{https://clemgris.github.io/I-FailSense/}{\textbf{Webpage}}).
\end{abstract}

\section{Introduction}
Humans and animals learn through failure \cite{thorndike1898animal}, a fundamental driver of intelligence and improvement. 
For robotic systems, the ability to automatically evaluate and learn from failure is critical to achieve robust general-purpose manipulation in open-world settings. 
Automatic failure evaluation has far-reaching implications: it enables reward shaping~\cite{ak2023learning}, improves motion planning~\cite{pan2022failure}, facilitates sub-task verification~\cite{ma2004validation}, and provides actionable feedback that enhances downstream robotic applications. 
This capability is particularly important for foundation models for robotics, such as \emph{Vision-Language-Action} (VLA) models \cite{driess2023palme, octomodelteam2024octo, kim2024openvla}, which train policy networks to infer actions from multi-modal input consisting of visual observations and language instructions. 
Within such foundation models, feedback-driven techniques, e.g. learning from human feedback \cite{xia2025robotic}, have shown that iterative feedback loops can guide and improve performance. 
However, a critical open question remains: \emph{How can robotic foundation models autonomously detect failures without relying on external supervision?}

\begin{figure}[!ht]
    \centering
    \includegraphics[width=1\linewidth]{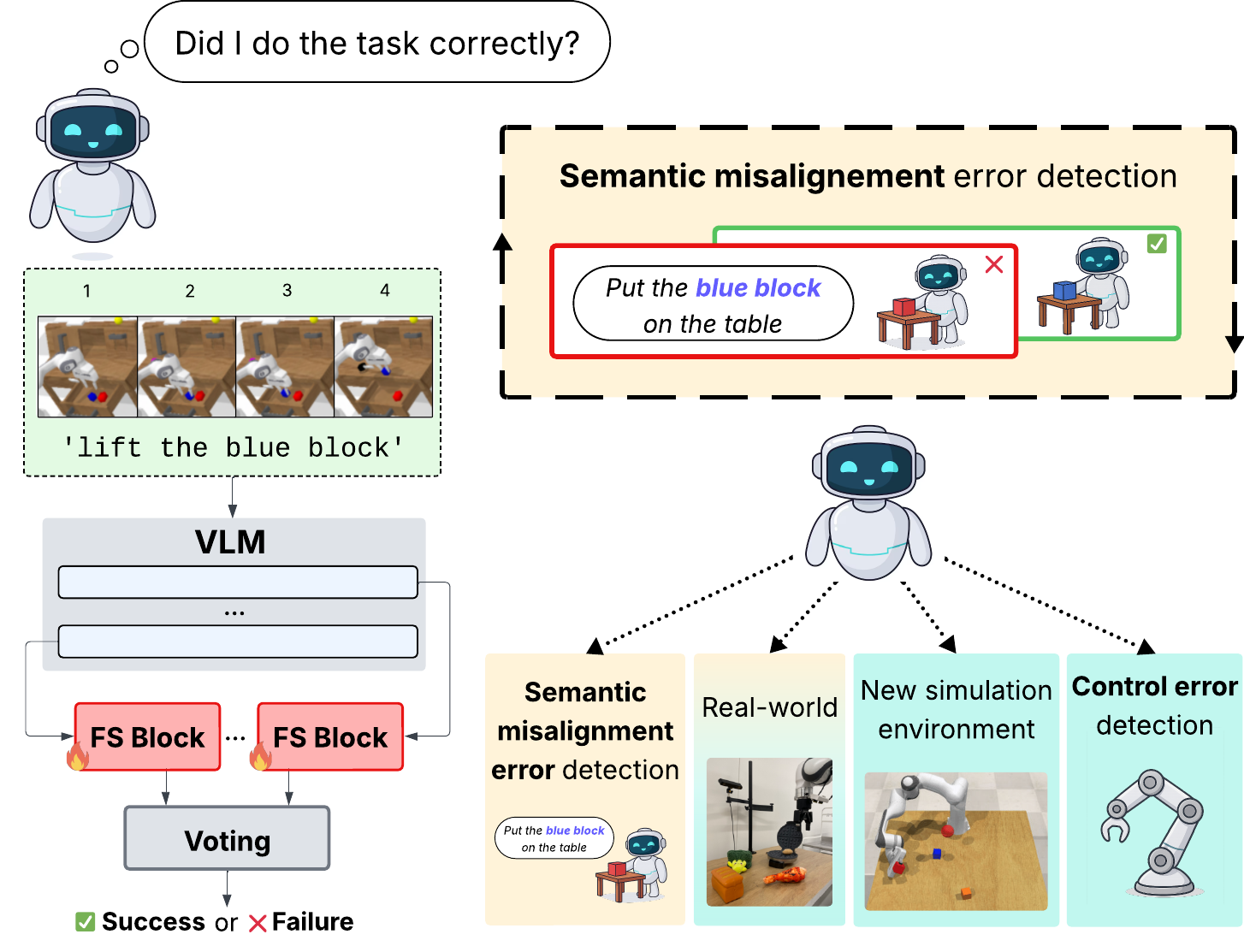}
    \caption{\textbf{Overview of I-FailSense}, which classifies a robot’s observation trajectories conditioned on language instructions into \textit{failure} or \textit{success}. Trained on semantic misalignment failure detection, I-FailSense excels at identifying these challenging errors, zero-shot generalizes to detecting control errors and errors in new simulation environments, and detects errors in real-world observations with minimal post-training.}
    \label{fig:intro}
\end{figure}

Recent advances in \emph{Vision-Language Models} (VLMs) highlight their potential as building blocks for robotic agents, as they excel at multi-modal open-world reasoning, showing strong capabilities in instruction following, prediction, and spatial reasoning across images, video, audio, and time series \cite{xu2025qwen2}, and are increasingly being applied in robotics for tasks such as object recognition, motion planning, spatial reasoning, and zero-shot trajectory generation \cite{shao2025large}.
However, despite these advances, the detection of failures in robotic trajectories remains an open problem for off-the-shell VLMs. Previous work has introduced benchmarks and fine-tuning strategies for VLM-based failure detection \cite{duan2024aha, gupta2025enhancing}, but these efforts primarily target \textbf{control errors}, e.g. the robot fails to grasp an object or drops it from the gripper. Far less attention has been given to \textbf{semantic misalignment errors}, where the robot demonstrates a semantically meaningful behavior but that mismatches the given instruction. However, these failures highlight a fundamental limitation of VLMs, and by extension of robotic agents built upon them, which is the lack of grounding between visual observations, motion, and semantic instructions \cite{yun2021does, tong2024eyes, peng2024finegrained, zhou2025vlm4d}.
From a failure detection perspective, these errors also raise an important challenge as, unlike control errors which can be more easily identified from visual clues, semantic misalignment errors require aligning the instruction with both the spatial dimension of object-level interactions and the temporal dimension of the robot’s movements. Addressing this challenge requires models that integrate language grounding with trajectory reasoning.
\medbreak
In this work, we introduce a new approach for the detection of language-conditioned robotic manipulation failure, with a focus on semantic misalignment errors, as shown in Figure~\ref{fig:intro}. To address this challenge, we build dedicated datasets from existing benchmarks and propose \emph{I-FailSense}, a failure-aware extension of VLMs. Our method employs a two-stage post-training pipeline: (1) parameter-efficient fine-tuning (PEFT) of the base VLM with LoRA, followed by (2) freezing the VLM and training lightweight binary classification heads for failure detection, called \emph{FailSense blocks} (FS blocks), over its internal language representations. Each block predicts the success or failure of the robot's behavior in the observed trajectory with respect to its given instruction, and the final prediction is obtained by combining the outputs of the blocks through an arbitration mechanism. This design explicitly grounds failure detection in language-space reasoning. 
\medbreak
Empirically, we demonstrate that I-FailSense outperforms zero-shot SOTA VLMs in detecting semantic misalignment errors reaching an accuracy of 90\% in our proposed datasets built from simulated trajectories. Remarkably, I-FailSense generalizes beyond its training data and achieves strong performance on the AHA dataset~\cite{duan2024aha}, which includes (1) unseen simulated environments and (2) unseen error category (control errors), even surpassing the VLM-based baseline trained directly on these tasks by +19 points in accuracy. Furthermore, I-FailSense transfers to real-world semantic misalignment failure detection with minimal fine-tuning of the FS blocks, achieving 74\% accuracy.
%

\begin{figure*}[!ht]
    \centering
    \includegraphics[width=1\linewidth]{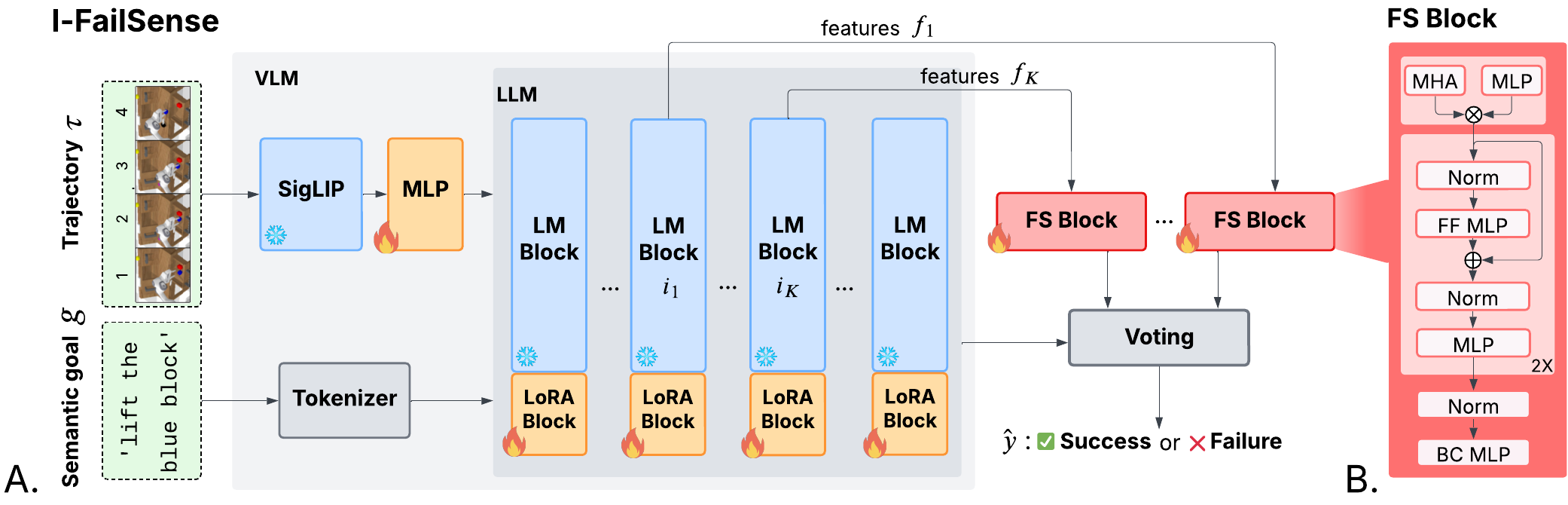}
    \caption{\textbf{I-FailSense Architecture.} (A) The model takes as input an observation trajectory aggregated into a single image $\tau \in \mathbb{R}^{3\times (H.N)\times(W.T)}$ (here, $N$=1 PoV and $T$=4 timesteps) and a semantic goal $g$, and outputs a binary success/failure prediction $\hat y$. I-FailSense is built on a base VLM (PaliGemma2-mix~\cite{steiner2024paligemma2}) and is post-trained in two stages: (1) the projection MLP is fine-tuned along with LoRA adapters applied to the language modules of the VLM's LLM base model, and (2) the VLM is frozen while the FS blocks, attached to the adapted language modules, are fine-tuned for binary classification. The FS block outputs are aggregated with the VLM’s final output through a voting mechanism to produce the final prediction. (B) The FS block architecture shows an hybrid attention pooling module composed of multi-head attention (MHA) and MLP followed by residual MLP blocks with batch normalization and ending in a binary classification MLP.}
    \label{fig:arch}
\end{figure*}

\section{Related Work}

\subsection{Failure Modes Analysis}
Failure detection and diagnosis are core problems in robotics. Early work in Human-Robot Interaction (HRI) and Task and Motion Planning (TAMP) \cite{garrett2020pddl} showed that transparent failure explanation is critical for trust in collaborative scenarios \cite{ye2019human,khanna2023user}. These studies highlight the importance of not only detecting but also communicating failures. From the point of view of researchers and engineers, understanding the failures of an agent means understanding its shortcomings and is therefore crucial to improving it. In fact, efforts have been made to classify failures into fixed \cite{duan2024aha} or learned \cite{gupta2025enhancing} interpretable failure modes (e.g., \textit{wrong rotation}, \textit{wrong grasp}, etc.). Recent work has also focused on building automated frameworks for failure analysis aimed at automatic policy improvement. Among these, AutoEval is a system for the autonomous evaluation of generalist robot policies that incorporates automatic success detection and scene resets, enabling scalable assessment in real-world settings \cite{zhou2025autoeval}. As another instance, the Robot Manipulation Diagnosis (RoboMD) framework employs deep reinforcement learning to actively explore and identify failure modes based on vision-language embeddings \cite{sagar2025robomd}. 
Our work aligns with this literature as it tackles generalizable failure detection and includes a reflection on failure modes analysis. Yet most existing studies primarily target control errors, which can be explained by incorrect motions of the robot and a lack of physical control or precision. In contrast, we focus on semantic misalignment errors, which differ from control errors in that they are not caused by low-level mistakes but by incorrect high-level understanding: 
They reveal that the agent fails to ground the given language instructions in its visual observations and motion.

\subsection{VLMs and LLMs for Failure Detection}
The integration of VLMs and LLMs has recently become a predominant paradigm for enhancing robotic manipulation, bringing a renewed focus to failure detection capabilities. A common approach has been to utilize off-the-shelf VLMs and LLMs as zero-shot or few-shot success detectors \cite{ma2022vip,ha2023scaling,wang2023gensim,skreta2024replan}. Some studies have adapted these models specifically for the task of failure detection through instruction-tuning \cite{du2023vision}. Building on this progress, AHA \cite{duan2024aha} represents a substantial advancement by formulating failure analysis as a generative, free-form reasoning task. This capability for rich explanatory reasoning facilitates integration with downstream systems for error recovery, leading to marked improvement in policy performance and demonstrating superiority over general-purpose models like GPT-4.
Our work, I-FailSense, builds directly upon the progress in VLM-based failure reasoning. While AHA powerfully demonstrated the value of generative explanations, its architecture is not explicitly optimized for the \textit{discriminative} task of high-accuracy failure classification—a prerequisite for triggering robust recovery mechanisms. In contrast, we decouple the explanatory power of a foundation VLM from a dedicated, trainable error classification head, ultimately achieving more accurate failure detection.

\subsection{VLM Representations for Multimodal Reasoning}

Failure detection, and especially semantic misalignment error detection, can be viewed as a challenging instance of the visual grounding problem: the model must verify whether a semantic goal has been achieved in a visual input composed of multiple scenes where a robot interacts with its environment. Pre-trained VLMs learn rich cross-modal representations during training \cite{luo2025visionlanguage} that can help address the visual grounding problem. However, although VLMs are trained on large internet-scale datasets for cross-modal tasks that require grounding textual instructions into visual inputs \cite{zhang2024visionlanguagemodels}, recent studies show that SOTA models still struggle with fine-grained visual grounding \cite{peng2024finegrained} and temporal/spatial reasoning about motion \cite{zhou2025vlm4d}. 
\medbreak
Under this perspective, prior works have explored enhancing VLMs by leveraging intermediate feature representations, often adding layers to extract and refine semantic information from middle layers \cite{tao2024probing}. These refined representations can then be used for downstream robotic monitoring tasks, such as automatic failure classification \cite{gupta2025enhancing} or detection \cite{gu2025safe}. Closer to our work, SAFE~\cite{gu2025safe} uses the features from the final layer of a VLA (built on a VLM) to predict control errors. In contrast, we exploit multiple levels of representation and aggregate predictions from those representations using a voting mechanism—a widely used strategy for improving both robustness and accuracy \cite{lin2025vote}--to detect control errors and semantic misalignment errors.

\section{Preliminary}

\subsection{Problem Statement}

The problem of robotic manipulation from language instruction and visual observations can be formulated as a goal-conditioned Partially Observable Markov Decision Process (POMDP) \cite{kaelbling1998pomdp}: $\mathcal{M} = (S, A, \mathcal{T}, \rho_0, \Omega, O, G),$ where $S$ denotes the state space, with the initial state sampled from the distribution $\rho_0$. The robot receives partial observations through the observation function $O$, which corresponds to its visual sensors. With $N$ camera point of views (PoV), the observation is represented as $O(s) \in \Omega = \mathbb{R}^{3 \times H \times W \times N}
$, where $H$ and $W$ are the image height and width. The robot selects actions in the action space $A$ (such as joint configurations or
\textit{n-DoF} commands) conditioned on observation $O(s)$ and a textual goal $g \in G \subseteq \mathcal{V}^*$, where $\mathcal{V}$ is a vocabulary and $\mathcal{V}^*$ all finite sequences over $\mathcal{V}$. The transition function $\mathcal{T}$ governs the environment dynamics.
Given this formulation, we define a robot trajectory of size $T$ as a sequence of observations, $\tau = (o_0 \dots, o_T)$. In this work, we are interested in the binary classification of trajectories into \textit{success} of \textit{failure} with respect to the textual goal $g$.

\subsection{Failure Modes in Robotic Manipulation}
\label{subsec:fail_modes}

Robotic failures can arise from different factors that reflect different shortcomings. In Duan et al. \cite{duan2024aha}, robotic manipulation failures are decomposed into seven categories. Among these, six are attributed to \textbf{control errors} (e.g., incomplete grasp, incorrect rotation, etc.), while only one (Wrong Target Object) corresponds to a misalignment between the language instruction and the action executed by the agent, e.g. the robot interacts with the wrong object. Yet, such semantic errors are not limited to wrong-object failures and can extend to broader forms of \textbf{semantic misalignment errors} where the agent behavior mismatches the language goal instruction. Whereas the other categories in AHA characterize low-level control mistakes, this type of failure highlights a lack of semantic grounding of the agent, a challenge that has been well documented in the literature \cite{yun2021does, tong2024eyes, peng2024finegrained}. Detecting such errors is therefore crucial for improving the grounding capabilities of robotic agents.
\medbreak
However, unlike control errors, which can often be detected without semantic reasoning (e.g., simply observing that the robot failed to grasp any object), identifying semantic misalignment errors requires deeper understanding of the relations between textual instructions and observation trajectories. As shown in the examples in Figure~\ref{fig:d_smf}, detecting such failures requires reasoning not only about the object being manipulated but also about the robot’s spatial movements over time, therefore requiring grounding the textual instruction in the spatial and temporal dimensions of the observation trajectory.

\section{Method}
\label{sec:method}

In this work, we focus on detecting language-conditioned robotic manipulation failures, with an emphasis on semantic misalignment errors. We introduce datasets targeting this challenging failure detection problem (Section~\ref{subsec:dataset}), and propose a new method for post-training a VLM agent and leveraging its internal representation for failure detection (Section~\ref{subsec:failsense}). Figure~\ref{fig:arch} presents the overall architecture.

\subsection{Semantic Misalignment Failure Datasets}
\label{subsec:dataset}

Our first contribution is a way to construct datasets $\mathcal{D}_{\text{SMF}}$ focusing on \textbf{S}emantic \textbf{M}isalignment \textbf{F}ailures. To this end, we leverage existing multi-task, language-conditioned robotic manipulation datasets that provide expert demonstrations $\{\tau, g^*\}$, where the semantic goal $g^*$ is successfully achieved in trajectory $\tau$. From any expert demonstration dataset, we define $\mathcal{D}_{\text{SMF}} = \{\tau, g, y\}$, where $\tau$ is the observation trajectory, $g$ the textual instruction, and $y$ a binary success label. Half of the examples are positive ($y=1$ with $g=g^*$), while the other half are negative ($y=0$ with $g \in G \setminus \{g^*\}$).

\medbreak
\textbf{Positive examples:} These correspond to the expert trajectories paired with their original textual instructions $\{\tau, g^*, y=1\}$.

\medbreak
\textbf{Negative examples:} To create negative examples for semantic misalignment, we take existing expert trajectories and pair them with textual instructions different from the original one, but belonging to the same task category (e.g., \textit{lifting}, \textit{pushing}, etc.). These negative examples are challenging in two ways: (1) the agent demonstrates behavior that is semantically meaningful but that does not correspond to the paired instruction, thereby characterizing semantic misalignment failures; (2) the paired instruction and the original one belong to the same task type, making the distinction subtle (e.g., "lift the red block" vs. "lift the blue block," or "rotate the red block right" vs. "rotate the red block left").

\subsection{FailSense: Vision-Language Model for Failure Detection}
\label{subsec:failsense}
Our second contribution is a post-training approach that adapts a pretrained VLM, PaliGemma2-mix-3B \cite{steiner2024paligemma2}, for robotic failure detection. We select this base VLM for its good trade-off between its size (3B parameters) and performance. Our method, illustrated in Figure~\ref{fig:arch}, follows a two-stage training: first, we post-train the VLM on the failure detection task using LoRA (\ref{subsec:ft}); second, we introduce our novel classification heads called \textit{FailSense (FS) blocks}, which are attached to multiple layers of the post-trained VLM and whose outputs are combined using an ensembling method for robust failure detection (Section~\ref{subsec:grounded_arbritation}). 

\medbreak
\subsubsection{Supervised Fine-Tuning (Stage 1)}
\label{subsec:ft}

We post-train the VLM by fine-tuning the MLP layer between the frozen vision encoder (SigLIP~\cite{zhai2023siglip}) and the base language model. Simultaneously, to achieve parameter-efficient adaptation of the language component, we incorporate Low-Rank Adaptation (LoRA)~\cite{hu2021lora} modules into the key-query-value (KQV) projection layers of attention blocks of the language model. This strategy introduces a small number of task-specific trainable parameters while keeping the majority of the pretrained model weights frozen.
\medbreak
The VLM processes a multi-modal input consisting of an observation trajectory \(\tau\) and a textual instruction \(g\). The trajectory \(\tau\) is aggregated into a single image representation $ \tau \in \mathbb{R}^{3 \times (H \cdot N) \times (W \cdot T)}$, where \(N\) denotes the number of point of views and \(T\) the number of observations. The combined input is formatted through a fixed prompt template $P(\tau,g)$, which is then fed into the model. 
\medbreak
For supervised training, we minimize the cross-entropy loss between the distribution predicted by the model and the target supervision token (\texttt{<success>} or \texttt{<fail>}):
\begin{equation}
\mathcal{L}_{\text{CE}}(\theta) = - \log p_\theta \left( t^* \mid x\right),
\end{equation}
where \(t^*\) is the target token, \(x\) is the input context (i.e. the prompt), and \(p_\theta(\cdot)\) denotes the probability distribution of the model with parameters \(\theta\), over the vocabulary.

\medbreak
\subsubsection{Grounded arbitration (Stage 2)}
\label{subsec:grounded_arbritation}
To leverage the diverse intermediate representations learned by the VLM for failure detection, we attach classification heads to \(K\) layers, indexed by \(\{i_k\}_{k=1}^K\), which are evenly distributed across the depth of the base language model. As illustrated in Figure~\ref{fig:arch}, each head takes as input features from a different depth of the model. At inference time, the outputs of all heads are aggregated to produce the final prediction, allowing binary classification that accounts for varying levels of reasoning abstraction. All classification heads are randomly initialized but share the same architecture. During this second training phase, only the FS blocks are updated. 

\medbreak
\textbf{FailSense blocks:}
The architecture of our classification heads is shown on the right of Figure~\ref{fig:arch}. Each head takes the features from its corresponding post-trained VLM layer $f_k=VLM_{\theta'}(P(\tau, g))[i_k]$ of the LLM block, and aggregates them using a hybrid attention pooling mechanism that combines MLP and Multi-Head Attention (MHA). The aggregated features are then processed through a sequence of MLP blocks with residual connections and batch normalization. A final binary classification linear layer predicts the binary probability $p_k = FS_{\phi_k}(f_k)$. At training time, $\phi_k$ is updated based on the binary cross entropy between the $k$-th head prediction $p_k$ and the true label $y$.

\medbreak
\textbf{Voting Strategy:}
At inference, each FS block outputs a probability distribution $p_k$, which is converted into a binary prediction $y_k = \arg\max(p_k)$. We also convert the free-form output of the VLM into a binary prediction $y_{vlm}$. The final prediction $\hat y$ is computed using a weighted voting mechanism:

\begin{equation*}
    \hat y = \mathds{1}\Bigg[\sum_{k=1}^{K} \omega_k \, y_k + \omega_{vlm}y_{vlm}> 0.5 \left(\sum_{k=1}^{K} \omega_k +\omega_{vlm}\right)\Bigg],
\end{equation*}
where $\omega_k$ denotes the weight of the $k$-th FS block and $\omega_{vlm}$ of the VLM prediction. In practice, we use $K=3$, assign equal weights to the FS blocks predictions ($\omega_k = 1$) and give the VLM prediction double weight ($\omega_{vlm} = 2$) to break ties. 

\section{Experimental Setup}

We train I-FailSense on a semantic misalignment dataset, built following the pipeline presented in Section~\ref{subsec:failsense}. We design experiments addressing the following research questions:
\textbf{Q1}: \textit{Can I-FailSense reliably detect semantic misalignment errors?} \textbf{Q2}: \textit{Does this capability transfer to detecting control errors?} \textbf{Q3}: \textit{How well does I-FailSense generalize to failure detection on out-of-distribution (OOD) simulated trajectories?} \textbf{Q4}: \textit{How well does it generalize to failure detection on real-world ?}


\medbreak
\noindent In addition, we conduct an ablation study to evaluate the contribution of each training stage in our pipeline to the overall performance.

\subsection{Datasets}
\label{subsec:eval_dataset}

\par
\begin{figure}[!ht]
    \centering
    \includegraphics[width=1.\linewidth]{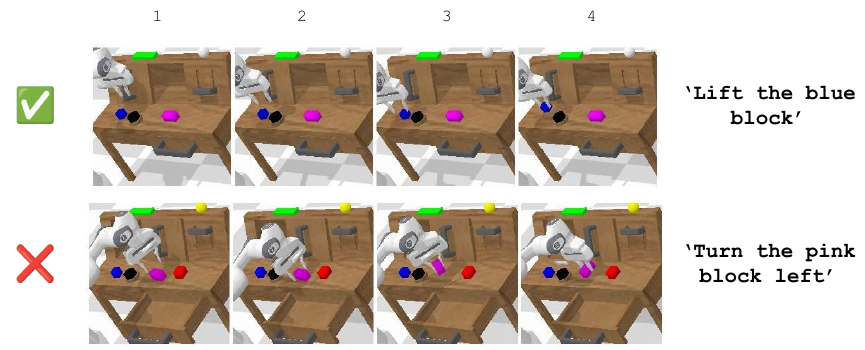}
    \caption{\textbf{Example data in $\mathcal{D}_{\text{SMF-CALVIN}}$}: Top: a positive example where the observation trajectory correctly matches the paired instruction. Bottom: a negative example illustrating semantic misalignment, where the robot rotates the correct object—the pink cube—right instead of the instructed left.}
    \label{fig:d_smf}
\end{figure}

 \textbf{SMF-CALVIN dataset:}
 To address $Q_1$, we build a semantic misalignment failure dataset $\mathcal{D}_{\text{SMF-CALVIN}}$ based on the CALVIN task\_D benchmark~\cite{mees2022calvin}. This benchmark comprises 34 simulated robotic manipulation tasks, which we grouped into six categories: \textit{lifting}, \textit{rotating}, \textit{pushing}, \textit{opening/closing}, \textit{placing}, and \textit{lighting}. For each task, the benchmark provides around 150 expert demonstrations collected via teleoperation and labeled with a textual instruction corresponding to the task. Each task includes 11 different textual instructions to increase semantic complexity. Figure~\ref{fig:d_smf} shows an example of a positive and a negative example.

\par
\begin{figure}[!h]
    \centering
    \includegraphics[width=1\linewidth]{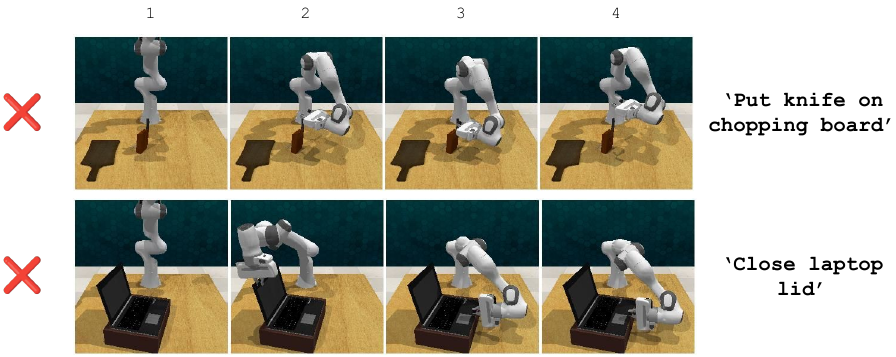}
    \caption{\textbf{Example data in $\mathcal{D}_{\text{AHA}}$}: Two negative examples from the AHA dataset (exocentric PoV) demonstrating control failures—top: the knife slips through the robot's gripper; bottom: the robot fails to grasp the computer lid.}
    \label{fig:AHA}
\end{figure}

 \textbf{AHA dataset:}
To address both $Q2$ and $Q3$, we evaluate I-FailSense trained on $\mathcal{D}_{\text{SMF-CALVIN}}$, on the $\mathcal{D}_{\text{AHA}}$ dataset~\cite{duan2024aha}. This failure detection benchmark is built on RLBench~\cite{james2020rlbench} and comprises 79 different simulated tasks. $\mathcal{D}_{\text{AHA}}$ defines seven error categories: six control errors (incomplete grasp, inadequate grip retention, misaligned keyframe, incorrect rotation, missing rotation, and wrong action sequence) and one semantic misalignment error (wrong target object). Failures in these categories are generated heuristically from RLBench as detailed in $AHA$~\cite{duan2024aha}. Observation trajectories include both exocentric and egocentric PoV. Since the dataset is not publicly released, we constructed a test set of 400 negative trajectory–instruction pairs following the $AHA$ pipeline; two representative examples (exocentric PoV) are illustrated in Figure~\ref{fig:AHA}.

\par
\begin{figure}[ht]
    \centering
    \includegraphics[width=\linewidth]{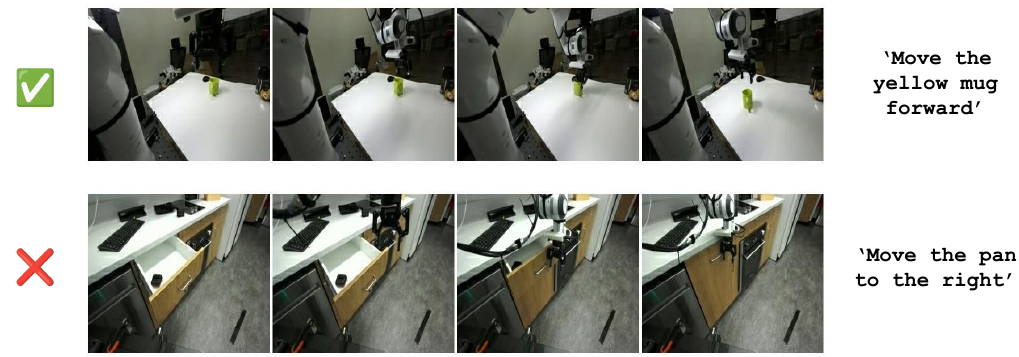}
    \caption{\textbf{Example data in $\mathcal{D}_{\text{SMF-DROID}}$}: Two examples from the semantic misalignment failure dataset built on DROID (exocentric PoV)--top: a positive example where the instruction matches the observation trajectory; bottom: a negative example where the instruction and trajectory mismatch.}
    \label{fig:droid}
\end{figure}

 \textbf{SMF-DROID dataset:}
To address $Q4$, we construct a second semantic misalignment failure dataset, $\mathcal{D}_{\text{SMF-DROID}}$, derived from the DROID dataset~\cite{khazatsky2025droid}. DROID consists of real-world robot demonstrations recorded from multiple PoV: one egocentric (gripper) and two exocentric cameras. In $\mathcal{D}_{\text{SMF-DROID}}$, these demonstrations serve as positive examples. To generate negative examples, we use the pipeline described in Section~\ref{subsec:dataset} and pair each trajectory in DROID with a randomly mismatched instruction, considering only one task category. The resulting dataset $\mathcal{D}_{\text{SMF-DROID}}$ contains $6K$ training and $276$ test data, with half positive and half negative trajectory–instruction pairs (see Figure~\ref{fig:droid}).

\subsection{Baselines and Metrics}

\textbf{Baselines:} We compare I-FailSense against several SOTA zero-shot VLMs from different model families, including models of comparable sizes (I-FailSense base model PaliGemma2-mix-3B) as well as larger ones (Qwen2.5-VL-7B~\cite{bai2025qwen25vl} and GPT-4o~\cite{openai2024gpt4o}). For evaluation on $\mathcal{D}_{\text{AHA}}$, we additionally report results of the AHA model \cite{duan2024aha}. The AHA model shares the same input format as I-FailSense, i.e. the observation trajectory flattened into a single image $\tau \in\mathbb{R}^{3\times(H.N)\times(W.T)}$. Its fine-tuning procedure also closely parallels the first training stage of I-FailSense: the model is initialized from LLaMA-2-13B, with fine-tuning applied only to the projector and transformer weights, while using a frozen vision encoder and tokenizer, following the pipeline introduced in \cite{liu2023visual}.

\medbreak
\textbf{Metrics:}
For evaluation on datasets composed of both positive and negative examples ($\mathcal{D}_{\text{SMF-CALVIN}}$ and $\mathcal{D}_{\text{SMF-DROID}}$), we use standard binary classification metrics, which are accuracy, precision (true positives over predicted positives), recall (true positives over real positives), and F1 score.
For evaluation on $\mathcal{D}_{\text{AHA}}$, which only contains negative examples, we report accuracy, which can be interpreted as a failure detection rate.

\section{Results}
\label{sec:results}

\subsection{Semantic Misalignment Failure Detection (Q1)}

To evaluate the performance of I-FailSense in detecting semantic misalignment failures, we tested it on the set of $\mathcal{D}_{\text{SMF-CALVIN}}$ tests. These data are in-distribution both in terms of error type (semantic misalignment failures) and environment (CALVIN). Table~\ref{tab:smf} presents the results of I-FailSense after one training phase, labeled \textit{(LoRA only)} and the full training, labeled \textit{(ours)}, compared to zero-shot VLMs. The results are shown when using one exocentric PoV ($N$=1) or an exocentric and egocentric PoV ($N$=2).

\begin{table}[h]
\footnotesize
\centering
\caption{Evaluation metrics of I-FailSense on $\mathcal{D}_{\text{SMF-CALVIN}}$ after Stage 1 \textit{(LoRA only)} and Stages 1 + 2 \textit{(ours)}, compared to zero-shot VLMs, for failure detection from observation trajectories captured from one (exocentric) or two PoV (exocentric and egocentric). Best results are shown in bold.}
\resizebox{0.48\textwidth}{!}{
\begin{tabular}{cc|cccc}
\toprule
\textbf{Method} & \textbf{\#PoV}                                  & \textbf{Accuracy} & \textbf{Precision} & \textbf{Recall} & \textbf{F1}\\ \midrule

\multirow{2}{*}{\shortstack{GPT-4o\\(zs)}} 
& 1 & 0.5764 & 0.8125 & 0.2453 & 0.3768 \\
& 2 & 0.6305 & 0.7719 & 0.4151 & 0.5399 \\ \midrule

\multirow{2}{*}{\shortstack{PaliGemma2-mix\\3B (zs)}} 
& 1 & 0.4729 & 0.4961 & 0.5943 & 0.5408 \\
& 2 & 0.5271 & 0.5329 & 0.7642 & 0.6279 \\ \midrule

\multirow{2}{*}{\shortstack{Qwen2.5-VL\\7B (zs)}} 
& 1 & 0.6798 & 0.6323 & 0.9245 & 0.7510 \\
& 2 & 0.6946 & 0.6507 & 0.8962 & 0.7540 \\ \midrule

\rowcolor{color}
\multirow{1}{*}{\shortstack{I-FailSense}} 
& 1 & 0.8571 & 0.8667 & 0.8585 & 0.8626\\
\rowcolor{color}
(LoRA only) & 2 & 0.8227 & 0.8365 & 0.8208 & 0.8286 \\ \midrule

\rowcolor{color}
\multirow{1}{*}{\shortstack{I-FailSense}}     
& 1 & \textbf{0.9064} & \textbf{0.8850} & \textbf{0.9434} & \textbf{0.9132}\\
\rowcolor{color}
(ours) & 2 & \textbf{0.8818} & \textbf{0.8596} & \textbf{0.9245} & \textbf{0.8909} \\ 
\bottomrule
\end{tabular}}
\label{tab:smf}
\end{table}

We can see in Table~\ref{tab:smf} that zero-shot GPT-4o and PaliGemma2-mix (I-FailSense's base model) fail to detect semantic misalignment errors from observation trajectories, performing close to random no matter they use only the exocentric PoV or both exocentric and egocentric PoV. Zero-shot Qwen2.5-VL performs better, reaching 68\% with a single PoV and 69\% with two PoV. I-FailSense outperforms all the zero-shot VLMs, achieving 90\% accuracy with a single PoV (exocentric) and 88\% with both PoV.
Qwen2.5-VL achieves a recall of $0.92$ and $0.87$, but only a precision of $0.63$ and $0.65$, respectively using one and two PoV. This suggests that while the model correctly identifies successful trajectories with respect to the instruction, it frequently misclassifies misaligned trajectories as successful (i.e., false positives). This behavior arises from the semantic misalignment failure dataset, which is designed with challenging negative examples where differences between the robot's behavior in the trajectory and the paired instruction are subtle (see Section~\ref{subsec:dataset}). While I-FailSense still exhibits this precision-recall gap, it is reduced to $0.06$ and $0.07$, resulting in F1 scores of $0.91$ and $0.89$, compared to $0.75$ for Qwen2.5-VL.


\subsection{Generalization Across Error Types (Q2)}

\begin{figure}
    \centering
    \includegraphics[width=1.02\linewidth]{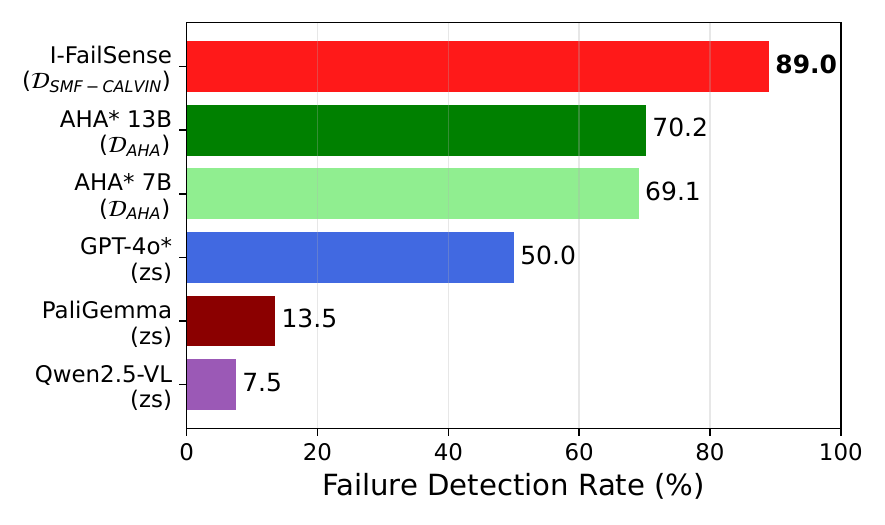}
    \caption{Failure detection rate of I-FailSense trained on $\mathcal{D}_{\text{SMF-CALVIN}}$\ and evaluated on $\mathcal{D}_{\text{AHA}}$, compared to $AHA$ baselines (7B, 13B) trained on $\mathcal{D}_{\text{AHA}}$ and zero-shot VLMs. All models use both egocentric and exocentric PoV. Results from prior work are marked with *.}
    \label{fig:aha_results}
    \vspace{-0.5cm}
\end{figure}

In Figure~\ref{fig:aha_results}, zero-shot VLMs perform poorly on failure detection in the $AHA$ dataset. GPT-4o achieves near-random accuracy, while PaliGemma2-mix-3B (base-model of I-FailSense) and Qwen2.5-VL perform even below random, likely due to the inherent bias of some VLMs for predicting success rather than failure and/or the simulated nature of the dataset. When fine-tuned on failure detection, the AHA models reach 69\% and 70\% accuracy for the 7B and 13B models. I-FailSense outperforms both, reaching $89\%$ accuracy, even though it was trained only on semantic misalignment errors, while $\mathcal{D}_{\text{AHA}}$ mostly contains control errors ($\approx$85\%) and only a small fraction ($\approx$15\%) of semantic misalignment errors (which is limited to wrong-object cases). This indicates that learning about semantic misalignment failures also provides strong signals to detect control errors. This can be explained by the fact that, when trained on semantic misalignment errors, the model learns to align language instructions with observation trajectories and to verify whether the robot’s motion matches the instruction (i.e. interacting with the right object in the right way). In misalignment errors, the robot performs a meaningful but incorrect motion (e.g. opening a drawer instead of the slider), while in control errors it fails to perform a meaningful motion at all (e.g. failing to grasp or dropping an object). Since both cases require checking whether the observed motion matches the instruction, training on semantic misalignment errors naturally transfers to control error detection.

\subsection{Generalization to OOD Simulations Trajectories (Q3)}

The strong performance of I-FailSense on the $AHA$ dataset (Figure~\ref{fig:aha_results}) demonstrates its ability to generalize across simulation environments. Specifically, $\mathcal{D}_{\text{AHA}}$ comprises trajectories from the RLBench environment \cite{james2020rlbench}, where a Franka arm interacts with varied objects (e.g., knifes, bowls, drawers etc.) relevant for the task, on a table (Figure~\ref{fig:AHA}). In contrast, I-FailSense was trained on $\mathcal{D}_{\text{SMF-CALVIN}}$, derived from the CALVIN environment \cite{mees2022calvin}, which also features a Franka arm (Figure~\ref{fig:d_smf}) but with a different and fixed set of objects (e.g., cube, lights, slider etc.). Furthermore, observations are captured from different camera settings. The good performance of I-FailSense on $\mathcal{D}_{\text{AHA}}$ indicates that the model has learned on $\mathcal{D}_{\text{SMF-CALVIN}}$ failure detection patterns that transfer effectively to other simulated environments.

\subsection{Generalization to Real-World Trajectories (Q4)}

\begin{table}[h]
\footnotesize
\centering
\caption{Evaluation metrics of I-FailSense on $\mathcal{D}_{\text{SMF-DROID}}$ with different fine-tuning of the FS blocks on $\mathcal{D}_{\text{SMF-CALVIN}}$ and/or $\mathcal{D}_{\text{SMF-DROID}}$ compared with zero-shot VLMs. FS blocks training dataset is into brackets, best results in bold.}
\resizebox{0.48\textwidth}{!}{
\begin{tabular}{cc|cccc}
\toprule
\textbf{Method} & \textbf{\#PoV}                                  & \textbf{Accuracy} & \textbf{Precision} & \textbf{Recall} & \textbf{F1}\\ \midrule

\multirow{2}{*}{\shortstack{GPT-4o\\(zs)}} 
& 1 & 0.5471 & \textbf{0.8421} &  0.1159 & 0.2038 \\
& 2 & 0.5797 & 0.8056 & 0.2101 & 0.3333 \\ \midrule

\multirow{2}{*}{\shortstack{PaliGemma2-mix\\3B (zs)}} 
& 1 & 0.5399& 0.5304 & \textbf{0.6957} & 0.6019 \\
& 2 & 0.5145 & 0.5099 & \textbf{0.7464} & 0.6059 \\ \midrule

\multirow{2}{*}{\shortstack{Qwen2.5-VL\\7B (zs)}} 
& 1 & {0.6884} & {0.8250} & 0.4783 & 0.6055 \\
& 2 & \textbf{0.7536} & \textbf{0.8646} & 0.6014 & 0.7094 \\ \midrule

\rowcolor{color}
\multirow{1}{*}{\shortstack{I-FailSense}}     
& 1 & 0.5580 & 0.6250 & 0.2899 & 0.3960\\
\rowcolor{color}
($\mathcal{D}_{\text{SMF-CALVIN}}$) & 2 & 0.6196 & 0.8000 & 0.3188 & 0.4560 \\  \midrule

\rowcolor{color}
\multirow{1}{*}{\shortstack{I-FailSense}}     
& 1 & 0.6594 & 0.7075 & 0.5435 & {0.6148} \\
\rowcolor{color}
($\mathcal{D}_{\text{SMF-CALVIN}}$ & 2 & 0.6848 & 0.7802 & 0.5145 & {0.6201}\\
\rowcolor{color}
  + $\mathcal{D}_{\text{SMF-DROID}}$) & & & & & \\ \midrule

\rowcolor{color}
\multirow{1}{*}{\shortstack{I-FailSense}}     
& 1 & \textbf{0.7100} & 0.7500 & {0.6304} & \textbf{0.6850} \\
\rowcolor{color}
( $\mathcal{D}_{\text{SMF-DROID}}$) & 2 & 0.7428 & 0.7680 & {0.6957} & \textbf{0.7300}\\ 
\bottomrule
\end{tabular}}
\label{tab:droid}
\end{table}

Table~\ref{tab:droid} shows the performance of I-FailSense evaluated on the $\mathcal{D}_{\text{SMF-DROID}}$ dataset, compared to zero-shot VLMs of different sizes. The I-FailSense models are trained with LoRA on $\mathcal{D}_{\text{SMF-CALVIN}}$, while the FS blocks are trained on $\mathcal{D}_{\text{SMF-CALVIN}}$ and/or $\mathcal{D}_{\text{SMF-DROID}}$. 
The results demonstrate that directly transferring I-FailSense fully trained on simulated trajectories ($\mathcal{D}_{\text{SMF-CALVIN}}$) to detect failures in real-world trajectories only partially succeeds: it improves the accuracy but decreases F1 score due to a significant drop in recall. This indicates that a model trained solely on simulated data struggles to recognize failures in real-world trajectories, even though precision increases, meaning that it better identifies successful trajectories. Training the FS blocks on both simulated and real-world trajectories  ($\mathcal{D}_{\text{SMF-CALVIN}}$ + $\mathcal{D}_{\text{SMF-DROID}}$) reduces the recall drop (from -0.4 to -0.23) while increasing precision ( +0.27), surpassing the base model in both accuracy and F1 score. The best performance is achieved by training the FS blocks solely on real-world failure detection ($\mathcal{D}_{\text{SMF-DROID}}$), suggesting that training on simulated data can hinder learning patterns relevant for real-world errors detection. Nevertheless, I-FailSense still benefits from the LoRA representation learned during the initial training, enabling effective failure detection in real-world trajectories. In fact, the top-performing I-FailSense matches the performance of the best zero-shot VLM, Qwen2.5-VL, despite having half the number of parameters.
Interestingly, while I-FailSense performs better using only the exocentric PoV compared to the dual-PoV model (+0.02 points in accuracy) for failure detection in simulated trajectories, on real-world trajectories the combination of exocentric and egocentric PoV achieves the best performance, surpassing the single-PoV model by +0.03 points in accuracy. This can be explained by the higher number of distractors and less stable viewpoints in real-world trajectories as depicted in Figure~\ref{fig:droid}, compared to the more controlled simulation environments (Figures~\ref{fig:d_smf} \ref{fig:AHA}). These factors make failure detection more challenging in real-world settings, hence the need for two PoV.

\section{Ablation Study}
\label{sec:ablation}

To study the impact of each training phase described in Section~\ref{sec:method}, we compare the base model with versions of I-FailSense trained in one or two phases. Table~\ref{tab:smf} reports the performance of the base model PaliGemma2-mix-3B, I-FailSense with LoRA adapters, labeled (\textit{LoRA} only), and I-FailSense with additional trained FS blocks (full training), labeled (\textit{ours}), when evaluated on $\mathcal{D}_{\text{SMF-CALVIN}}$. While zero-shot PaliGemma2-mix performs close to random in detecting semantic misalignment errors, with accuracies of $47\%$ and $52\%$, LoRA fine-tuning increases the model's performance to $85\%$ and $82\%$ (respectively for one and two PoV). Two factors can explain the base model’s poor performance and the need for fine-tuning. First, PaliGemma2-mix was trained on vision-language tasks (including image captioning, visual question answering, and multi-modal understanding) using real-world images, whereas failure detection in $\mathcal{D}_{\text{SMF-CALVIN}}$ is composed of trajectories in a simulated environment. Second, PaliGemma2-mix was trained mainly on static image while $\mathcal{D}_{\text{SMF-CALVIN}}$ requires grounding the textual instruction with an input image that represents the movement of a robot in a scene, which is a more challenging task (Section~\ref{subsec:fail_modes}).
\medbreak
The FS blocks are connected to different levels of the VLM’s LLM base model, meaning each block leverages a distinct level of reasoning abstraction to generate its prediction. The ensembling mechanism then combines these predictions, ensuring that all levels of representation contribute to the final decision.
As a result, adding these blocks (trained during the second phase) improves the model’s performance in both accuracy ($85\%\rightarrow90\%$ for one PoV and $82\%\rightarrow88\%$ for two PoV) and F1 score ($0.86\rightarrow0.91$ for one PoV and $0.82\rightarrow0.89$ for two PoV). This shows that using the VLM’s representations at different levels rather than using only the final outputs helps the model better understand the multi-modal inputs, ultimately improving overall performance.


\section{Conclusion}
\label{sec:conclusion}

In this work, we addressed the problem of detecting language-conditioned robotic manipulation failures, with a particular focus on semantic misalignment errors. We proposed a pipeline for creating datasets targeting this challenging failure detection problem from existing benchmarks with expert demonstrations. We also introduced I-FailSense, a two-stage framework which first post-trains a base VLM with LoRA and then trains classification heads (FailSense blocks) attached to different levels of the internal representations of the VLM and whose outputs are combined via arbitration to detect failures from observation trajectory. Through experiments on simulation and real-world datasets, we demonstrated that I-FailSense not only detects semantic misalignment errors, but also generalizes to the detection of control errors and novel simulation environments. Moreover, minimal fine-tuning enabled sim-to-real transfer. This shows that training on semantic failure detection, combined with an architecture that explicitly leverages different levels of representation of pre-trained VLMs, allows strong failure detection and generalization beyond training conditions.
I-FailSense represents a step toward language-conditioned agents that evaluate their own behavior. Future work will extend this framework to failure recovery, enabling robots to both recognize and adapt to errors during execution.

\small{
\section*{Acknowledgments}
 \noindent This work used IDRIS HPC resources under the allocation
2025-[AD011015740] made by GENCI, was
supported by the European Commission’s Horizon Europe Framework Programme under grant No 101070381
(PILLAR-robots), and was partially funded by the French National Research Agency (ANR) under the OSTENSIVE projects (ANR-24-CE33-6907-01), ANITA
project (ANR-22-CE38-0012-01) and the France 2030 program, reference ANR-23-PAVH-0005 (INNOVCARE Project).}

\bibliographystyle{IEEEtran}
\bibliography{references}

\end{document}